\documentclass[journal]{IEEEtran}
\usepackage[cmex10,tbtags]{amsmath}
\usepackage{graphicx,algorithm,algpseudocode,algorithmicx,epstopdf,multirow,amsmath,bm,fixltx2e,stackengine,booktabs,threeparttable,cite,verbatim,caption,color}
\usepackage{makecell,pbox}
\usepackage{marginnote}

\usepackage{float}
\newcommand\blfootnote[1]{%
	\begingroup
	\renewcommand\thefootnote{}\footnote{#1}%
	\addtocounter{footnote}{-1}%
	\endgroup
}

\begin{document}
		
\title{Enhanced Center Coding for Cell Detection with Convolutional Neural Networks}

\author{\IEEEauthorblockN{Haoyi Liang, \textit{Student Member, IEEE},  Aijaz Naik, Cedric L.  Williams, Jaideep Kapur, and Daniel S. Weller, \textit{Member, IEEE}}}
\maketitle

\begin{abstract}	
	Cell imaging and analysis are fundamental to biomedical research because cells are the basic functional units of life. Among different cell-related analysis, cell counting and detection are widely used. 
	%Recent research demonstrates that the convolutional neural network is a promising technique for object detection, and many state-of-the-art network architectures are adopted for cell counting and detection. 
	In this paper, we focus on one common step of learning-based cell counting approaches: coding the raw dot labels into more suitable maps for learning. Two criteria of coding raw dot labels are discussed, and a new coding scheme is proposed in this paper. The two criteria measure how easy it is to train the model with a coding scheme, and how robust the recovered raw dot labels are when predicting. The most compelling advantage of the proposed coding scheme is the ability to distinguish neighboring cells in crowded regions. Cell counting and detection experiments are conducted for five coding schemes on four types of cells and two network architectures. The proposed coding scheme improves the counting accuracy versus the widely-used Gaussian and rectangle kernels up to 12\%, and also improves the detection accuracy versus the common proximity coding up to 14\%. 
	
\end{abstract}

\textit{\textbf{Index Terms}}-- Cell counting, Cell detection, Convolutional neural networks

\section{Introduction}
\label{sec:intro}
	\begin{figure}[ht]
		\begin{minipage}[t]{0.99\linewidth}
			\centering
			\includegraphics[width = 0.99\linewidth, trim={0cm, 0.4cm, 21.7cm, 0cm}, clip]{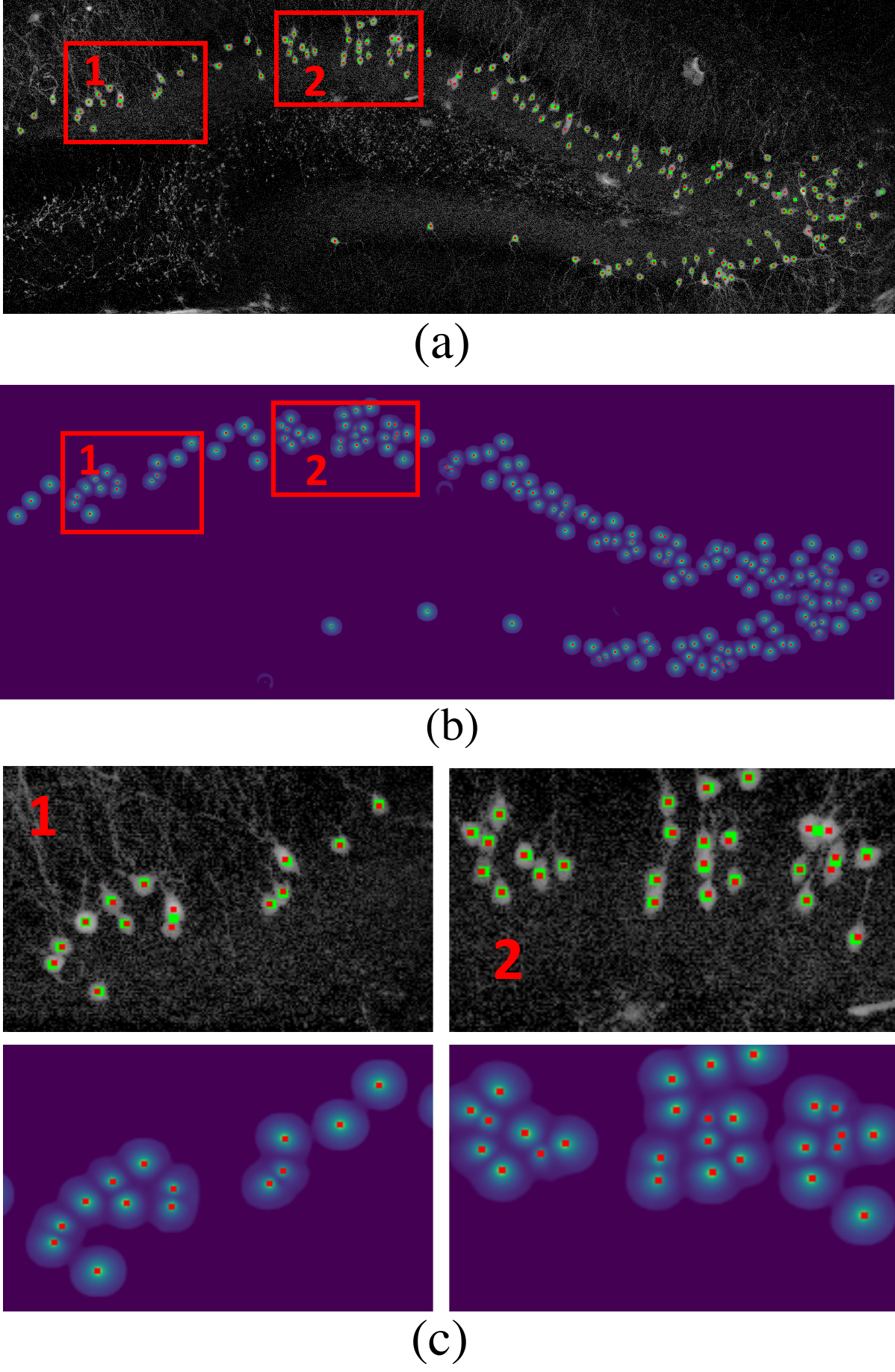}				
		\end{minipage}					
		\caption{Cell detection with the proposed repel coding for the granule cells in mouse brain tissues. (a) Input image. (b) Output of the network. (c) Zoomed-in patches. Green dots are the labeled cell centers, and red dots are the detected cell centers.}
		\label{fig:detection_overview}
	\end{figure}
	%How to automatically process the microscopy imaging data is a focus of biomedical research \cite{cell_seg_review}.
	\blfootnote{
	This work has been submitted to the IEEE for possible publication. Copyright may be transferred without notice, after which this version may no longer be accessible.
		
		Dr. Daniel S. Weller is supported by NSF 1759802.  Dr. Cedric L.  Williams is supported by ARO W911NF-16-C-0104. One Titan X Pascal GPU used in this work is supported by NVIDIA.
		
	H. Liang and D. S. Weller are with the Department of Electrical and Computer Engineering, University of Virginia,  Charlottesville, VA 22904 USA (email: hl2uc@virginia.edu; dsw8c@virginia.edu). A. Naik and J. Kapur are with the Department of Neurology, University of Virginia, Charlottesville, VA 22904 USA (email: an8hg@virginia.edu; jk8t@virginia.edu).		
	C. L.  Williams is with the Department of Psychology, University of Virginia, Charlottesville, VA 22904 USA (email: clw3b@virginia.edu).}

	The cell number and distribution pattern reflect many underlying biomedical mechanisms. For example, the cell number is important for the diagnosis and treatment of breast cancer\cite{breast_cancer}, and subcellular object counting and localization facilitate large scale tumor study with tissue microarrays\cite{tissue_array}.	
	With the advances of microscopy imaging in recent years, automatically extracting meaningful information from large amounts of raw imaging data is necessary in many applications. 	
	Cell counting and detection methods can be classified as feature-based methods and learning-based methods\cite{cell_seg_review}. Feature-based methods \cite{bright_field_cell, non_overlapping_extremal_regions, pop_out_cells} are  built with hand-designed features and parametric models. Common hand-designed features include  maximally stable extremal regions (MSER) \cite{MSER}, scale invariant feature transform (SIFT) \cite{SIFT} and histogram of gradients (HOG)\cite{HOG}. Level sets \cite{level_set} and active contours\cite{active_contour} are two widely used parametric models. Features based on local convexity are processed by a graph model for cell detection\cite{non_overlapping_extremal_regions}. Support vector machines (SVM) are fed with regional proposals for cell detection\cite{bright_field_cell, non_overlapping_extremal_regions}.  
	However, one limitation of feature-based methods is that the performance varies among different types of cells, and sometimes these features need to be redesigned for different cell targets.  
	Recently, data-driven methods receive more attention \cite{Unet, xie_16,yao_16} for two reasons. First, the success of deep learning in the area of object classification and detection\cite{alex_net} provides practical experience on the architecture design and training of convolutional neural networks (CNN) that can be adapted to other domains. Packages such as Pytorch \cite{pytorch}, and regularization techniques such as dropout\cite{drop_out} and residual convolution\cite{residual_net} make building customized CNNs easier. Secondly, huge amounts of imaging data are generated with advanced microscopy technology. Models trained with more imaging data are more robust and accurate.
	
	Because microscopy data differ from natural scene images in many ways, transferring solutions for natural scene images to cell analysis should be performed carefully.
	For cell analysis, the objects of interest, cells, usually are of a smaller size compared with the objects in natural scene images. %The first property of cell detection with microscopy data is that the object size is small. Compared with the tasks on natural scene image, the variance of cells is significantly smaller. Without considering the locality of pixel intensity, the object variance is exponentially related to the number of pixels. 
	Very deep network architectures are not necessary for cell detection\cite{yao_16}, since the typical size of a single cell is within $50\times50$ pixels in microscopy images. The benefit of designing a deeper network is marginal if the receptive field is large enough to cover a single cell. Deeper networks also have more trainable parameters, and this could even hurt the performance by over-fitting. 
	In biomedical research, cell analysis is a highly customized task. The morphology could change dramatically among different types of cells. Even for the same type of cell, the appearance changes with different tissue preparation protocols, imaging equipment and imaging protocols. Instead of training an omnipotent cell analysis framework, the ability to adapt the analysis model to a specific cell type and imaging modality is desirable in practice. 

	%However, for the same reason mentioned before, the requirement of big data can also be a limitation to learning-based method. For rare cell detection, where the number of training examples are very limited, the performance of learning-based method is deteriorated. Many techniques are proposed to tackle this challenges. Max out\cite{max_out} and drop out \cite{drop_out} avoid over-fitting to training data by weight selection during the backwards gradients descent. Residual learning \cite{residual_net} resolves the challenges caused by lack of data when the neural networks are very deep (over 100 layers). 

	The success of deep learning on numerous computer vision tasks is widely perceived as a result of improved learning architectures and big data. CNNs and many regularization techniques, such as max out\cite{max_out}, drop out \cite{drop_out} and residual convolution \cite{residual_net}, are practical components for building neural networks. 
	%With simple algorithm, if the data is larger enough, we can get satisfying result. 
	The other factor, data enhancement, is discussed less especially for the output data. 
	In this paper, instead of focusing on the data enhancement at the input side of a learning network, the coding scheme at the output side is discussed. Compared with the input data, the output data has more freedom to be designed to suit our applications. 
	For cell analysis, one type of the raw labels is a sparse 2D picture where non-zero pixels indicate centers of cells. The widely used cell density function for cell counting can be taken as an augmentation of the output data\cite{count_ception}. The cell density usually is a smoothed version of the raw dot labels. Similarly, cropping the whole images into patches can be taken as filtering the raw dot labels with a rectangle filter.
	
	To better understand the effect of output data enhancement, two criteria of raw label coding are discussed in this paper: entropy and reversibility. The entropy measures how suitable the coding scheme is for training, and the reversibility measures how robust inverting the coding to raw labels is when predicting. Different raw label coding methods are trading-off between these two criteria. For applications with different variations of shape, size and crowdedness of the cells, the coding scheme should be designed accordingly. %The proposed coding criteria provide a guideline to improve the cell detection performance with output coding. 
	The coding scheme with both high entropy and high reversibility is always preferred. However, a trade-off between these two indexes has to be made in most cases depending on the training loss and the overall performance. For example, if the training loss with a coding scheme converges fast, but the accuracy is not ideal after inversing the outputs to the raw data, then a coding scheme with lower entropy but higher reversibility should be considered. We also propose a new coding scheme that balances these two criteria well in most cases. Fig. \ref{fig:detection_overview} shows one example of granule cell detection in the mouse brain with the proposed coding scheme. However, it should be emphasized that we are not proposing an optimal coding scheme for all types of cells. The coding scheme should be carefully designed based on each application. 
	%In this paper, we hope the proposed criteria provide practical directions to explore when designing the coding of raw ground truth labeling.

	The rest of this paper is organized as follows. Section \ref{sec:related_work} reviews recent cell counting and detection works with CNNs, and highlights one common step in these works: raw dot label enhancement. Section \ref{sec:proposed_work} details the proposed coding evaluation criteria, and provides a cell center coding scheme based on the design criteria. Section \ref{sec:expriment} verifies the proposed coding method with four cell datasets and two network architectures. At last, Section \ref{sec:conclusion} reviews the proposed method, and discusses future work in cell analysis.

\section{Related works}
\label{sec:related_work}

	\subsection{Object Detection in Natural Scenes and Cell Analysis}	
	Object detection for natural scene images usually outputs a group of bounding boxes to represent the location and the size of detected objects. Both R-CNN\cite{R-CNN}, YOLO\cite{YOLO} and their variants adopt this box representation. Some widely used detection benchmarks also use the box labeling, such OTB\cite{tracking_benchmark} and COCO\cite{COCO}. By representing an object with a bounding box by four numbers, the number of output nodes in a detection network is significantly reduced. 
	For cell analysis, the center dots and contours of cells are more common labeling formats rather the bounding box. Since labeling contours for training costs more time, dot labeling is used more often if the cell morphology is not of particular importance. In this paper, we focus on the dot labeling for cell counting and detection.
	
	Among the learning-based cell detection works, the raw dot labels usually are pre-processed before being set as the output data for training. Cropping whole images into small patches\cite{yao_16,yuanpu_15}, and transforming the dot labels into a density representation \cite{count_ception,learn_to_count_with_boosting} are two common approaches.	
	
	Designed for cell counting\cite{yao_16}, the inputs of the CNN are patches of size $60 \times 60$ pixels, and the output is the total number of cells in this patch. The CNN treats number estimation as a regression problem rather than a classification one. The reason is that the appearance difference between patches with similar number of cells should be smaller than that between patches with great disparity in cell numbers\cite{yao_16}. By cropping a whole image into multiple patches, the quantity of training examples is greatly enhanced. With the implementation of fully convolutional networks \cite{FCN_semantic,Mask_RCNN}, another way to understand the pipeline of cropping and counting is that the raw dot labels are filtered with a moving average filter of size $60 \times 60$. 
	%The cropping approach is discussed in other literatures \cite{count_ception,learn_to_count_with_boosting}. The raw dot labels are density-coded by smoothing with a Gaussian kernel\cite{count_ception}. The input patch is of size $32 \times 32$, and the output is a single value indicating the cell number\cite{count_ception}. Similarly, with enough GPU memory, the cropping step\cite{count_ception} can be bypassed with fully convolutional implementation\cite{learn_to_count_with_boosting}. 
	In Walach's work \cite{learn_to_count_with_boosting}, raw dot labels convolved with a Gaussian kernel are used for cell detection, and raw dot labels convolved with a human-shape kernel are used for pedestrian detection.
	
	Cell counting and detection are implemented by a CNN followed by a compressive sensing module\cite{compressive_sensing_counting}. The input to the CNN is a patch of size $200 \times 200$ containing multiple cells, and the output is a vector $y$ that contains compressed center location information. The compressed location information is computed by a learned sensing matrix $S$ with $y=Sx$, where $x$ is a vector of raw dot labels. The length of $y$ is much less than the length of $x$. As a result, decoding with a sparse prior is used to recover the exact position at prediction. The concept of coding the raw dot labels is emphasized \cite{compressive_sensing_counting}, but the coding with compressive sensing is used less than simple spatial filter codings for two reasons. First, the sensing matrix, $S$, is an extra mapping relation learned by the neural network. Because of the size of $S$, a large amount of training data  is required to avoid over-fitting and maintain good accuracy. Second, recovering the location information from the compressed vector can be time-consuming. 
	
	Unlike coding the raw dot labels by convolving with a filter \cite{yao_16, count_ception},  proximity coding\cite{yuanpu_15} is defined as,
	\begin{equation}
	\label{eqn:proximity}
	C_{ij}=\begin{cases}
	\frac{1}{1+\alpha D_{ij}}, & \text{if $D_{ij}<r$},\\
	0, & \text{otherwise},
	\end{cases}
	\end{equation}
	\noindent{where $D_{ij}$ is the Euclidean distance from the pixel $(i,j)$ to the closest cell center. Because  proximity coding preserves local maxima at  cell centers, such coding can be used for cell detection as well \cite{xie_16}.}
	
	From recent works on cell analysis, there are two observations. First, transformation of the raw dot labels is necessary before training the network. As a result, a corresponding inverse transformation is required at the prediction phase. For cell counting, integration over the outputs serves this purpose, and for cell detection, local maximum detection is used. Second, cell counting and cell detection share many common components. The tasks of cell counting and detection share the same pipeline \cite{xie_16}. A more detailed comparison between cell counting and detection is provided in the next part.

	\subsection{Counting vs. Detection}
	Object counting usually is preferred over detection in applications where objects are crowded and single objects are not distinguishable. In this case, detection will not provide accurate location information, and texture information is a crucial clue to estimate the object density\cite{dense_counting_1, dense_counting_2}. However, if each single object can be identified, the detection approach has more advantages. First, detection provides the location information lost in counting. Second, more complexity is involved in counting than detection for training. For the applications of cell analysis, usually the size of cells is much smaller than the whole image. This indicates that a single cell can be  recognized with a smaller receptive field than the whole image required by counting. Larger receptive fields of CNNs usually mean more trainable parameters and deeper networks. As mentioned before, cell analysis is a highly customized application, and retraining a model to fit a specific imaging modality and cell morphology  is more effective than an overly general model. For this reason, cell detection enables a smaller network design that is easier to train with limited labeled data. In general, detection is a more appropriate approach if objects are not heavily occluded and each single object is recognizable. The major challenge of detection is when objects are densely packed or partially occluded. In the next section, two criteria are proposed for the raw dot label coding for cell detection, and a new raw dot coding method, repel coding, is proposed to better tackle this challenges.
	%If the patches size or the blurring kernel if small enough, the counting result can directly used for detection\cite{compressive_sensing_counting,xie_16}. 
	
\section{Proposed coding scheme}
\label{sec:proposed_work}
	\begin{table*}[h!t]
		\centering
		\caption{Entropy and Reversibility of Different Coding Schemes}
		\begin{tabular}{ | c | c| c | c | c | c | }
			\hline
			& Dot labels & Gaussian kernel &  Avg. kernel & Proximity coding & \textbf{Repel coding}    \\ \hline
			$E$ & $0.0000$ & $6.761$ & $2.052 \times 10^{-1}$ & $4.901$ & $6.472$  \\ \hline
			$R$ & $1.000$ & $4.583\times10^{-3}$ & $4.434\times10^{-3}$ & $1.198\times10^{-2}$& $8.199\times10^{-3}$  \\ \hline
			$R^5$ & $1.000$ & $1.089\times10^{-1}$ & $1.111\times10^{-1}$ & $1.286\times10^{-1}$ & $1.306\times10^{-1}$   \\ \hline
		\end{tabular}
		\label{tabel:coding_compare}
	\end{table*}
	\begin{figure}[t]
		\begin{minipage}[t]{0.99\linewidth}
			\centering
			\includegraphics[width = 0.99\linewidth, trim={0cm, 2.5cm, 21cm, 0cm}, clip]{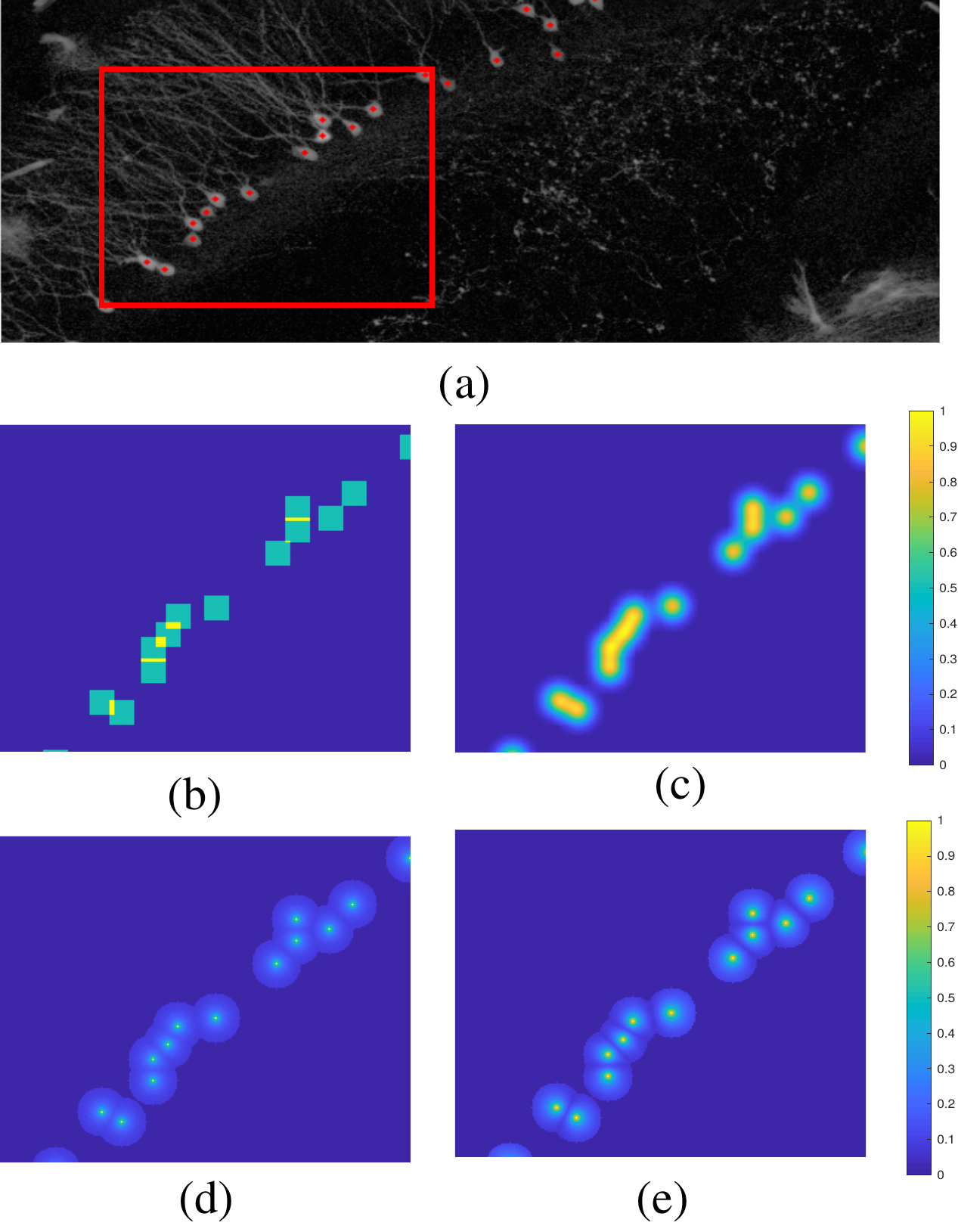}				
		\end{minipage}					
		\caption{Illustrations of different coding schemes. (a) An example image with cell centers labeled with red dots. The rectangle in (a) indicates the zoomed-in patches in (b)-(e). (b) shows the coding with the rectangle kernel\cite{count_ception}. (c) shows the coding with the Gaussian kernel\cite{count_ception,learn_to_count_with_boosting}. (d) shows the coding with  proximity coding\cite{xie_16,yuanpu_15}. (e) is the proposed repel coding. The proposed center coding scheme is superior to  proximity coding in two ways. First, the repel coding provides a slower response decay away from the cell center, which is helpful to the stability during training. Secondly, the repel coding suppresses the responses between two cell centers. This is important to recover the raw dot labels during prediction. Coding values in (b) to (e) are normalized to 0-1 for comparison.}
		\label{fig:coding_overview}
	\end{figure}
	
\subsection{Criteria}
	The two proposed criteria for the center coding are entropy and reversibility. At the training phase, entropy characterizes if the coding scheme is easy for the neural network to learn. At the prediction phase, reversibility measures if the coding scheme can recover the raw dot labels robustly.
	\paragraph{Entropy}
	The entropy of a coding scheme, $C$, is defined as，
	\begin{equation}
	\label{eqn:entropy}
	E_C = entropy(C_{ij} \ if\ C_{ij} \ne 0),
	\end{equation}

	\noindent{where $C_{ij}$ represents the coded value at position $(i,j)$. Zero values in the transformed coding are excluded here, since these regions usually are far from any cell. 
	Entropy measures how evenly the non-zero values are distributed.
	An ideal coding scheme should distribute the coding values uniformly over a range. By doing so, the gradient backpropagation during the training phase is more robust. A similar concept is mentioned by modeling the counting problem as regression rather than classification \cite{yao_16}. The extreme case of low entropy coding is the raw dot labels, where the entropy is always zero.}
	
	\paragraph{Reversibility}
	The reversibility of a coding $C$ is defined as,
	\begin{equation}
	\label{eqn:reversibility}
	R_C = \frac{\sum_{i,j} M_{ij}\cdot C_{ij}}{\sum_{i,j } C_{ij}},
	\end{equation}
	\noindent{where $M$ is the mask defining the proximity region of cell centers. The binarized raw dot labels, or the dilated version of raw dot labels can be used for $M$. In the prediction phase, the output response is not identical to the ideal coding scheme. A robust coding scheme should be able to recover the original coding, raw dot labels, in challenging cases. Reversibility is a similarity measurement between the raw dot labels and the coded response. Because local maximum detection is used to recover the raw dot labels at the prediction phase, reversibility here is defined as the degree of energy concentration around the raw dot labels. 
	
	For cell detection, a coding scheme with large entropy and reversibility indexes is preferred. As an extreme case,  the dot label itself has the maximum reversibility index. However, the raw dot label has the smallest entropy index. This means it is hard for the neural network to learn raw dot labels. On the other hand, coding by the Gaussian kernel has a larger entropy but a lower reversibility index. The result is that networks trained with coding by a Gaussian kernel converge fast and robustly in terms of loss value, but center recovery is obscured in the prediction phase. %In order to detect the cell center, a coding scheme that achieves a good balance between reversibility and entropy is desire. Next, we proposed our repel-distance coding. 
	More analysis on the entropy and reversibility trade-off is illustrated with experimental results in Section \ref{sec:expriment}.}

\subsection{Repel Coding}
	The proposed coding scheme of raw dot labels is based on  proximity coding defined in Eqn. \ref{eqn:proximity}. When  proximity coding was first proposed \cite{yuanpu_15}, it was designed for cell counting. Because  proximity coding produces local maxima at cell centers, it was also used for cell detection later \cite{xie_16}. However, one common challenge for cell detection is to distinguish two neighboring cells. For cell counting, only a global counting number is required. In other words, only the entropy is considered when coding raw dot labels for cell counting, but not the reversibility. In practice, we notice that  proximity coding does not perform well for detection when cells are crowded. The response valley between two cells is not significant enough, and local maxima do not align with cell centers accurately during the prediction. Aiming at increasing the reversibility of  proximity coding, the proposed repel coding is defined as,  
	\begin{equation}
	\begin{split}
	\begin{aligned}	
	&D'_{ij}=dist^1_{ij}\times {(1+dist^1_{ij}/dist^2_{ij})}^2,&	\\
	&C_{ij}=\begin{cases}
	\frac{1}{1+\alpha D'_{ij}}, & \text{if $D'_{ij}<r$},\\
	0, & \text{otherwise},
	\end{cases}&
	\end{aligned}
	\end{split}
	\end{equation}	
	\noindent{where $dist^1_{ij}$ is the distance of the pixel $(i,j)$ to its nearest cell center, and $dist^2_{ij}$ is the distance of the pixel $(i,j)$ to its second nearest cell center. The intermediate variable $D'_{ij}$ can be taken as $dist^1_{ij}$ suppressed by $dist^2_{ij}$.}

	In Fig. \ref{fig:coding_overview}, examples of different coding schemes are illustrated. Comparing Fig. \ref{fig:coding_overview} (d) and Fig. \ref{fig:coding_overview} (e), it is obvious that the proposed repel coding forms a more significant valley between two neighboring cells than  proximity coding. Table \ref{tabel:coding_compare} provides the entropy and the reversibility of different coding schemes shown in Fig. \ref{fig:coding_overview}. The entropy, $E$ in Table \ref{tabel:coding_compare}, is calculated by separating the coded non-zero values into eight bins. 
	The reversibility, $R$ in Table \ref{tabel:coding_compare}, is calculated by using the raw dot labels as $M$ in Eqn. \ref{eqn:reversibility}. Because it is rare in practice that the centers of two cells are 1 pixel away, the dilated reversibility, $R^5$ in Table \ref{tabel:coding_compare}, is calculated by dilating the raw dot labels with a disk of diameter of $5$ pixels. The meaning of the entropy and the reversibility in Table \ref{tabel:coding_compare} can be interpreted by comparing with the illustrations in Fig. \ref{fig:coding_overview}. The coding scheme with the highest entropy is the Gaussian kernel, and the entropy of the proposed repel coding is slightly less than that of the Gaussian kernel. With visual inspection, the intensity variations of the Gaussian kernel and the repel coding are larger than the other codings. Measured by $R^5$, the proposed repel coding achieves the highest reversibility index except for raw dot labels. This also aligns with the visual inspection where cell centers with the repel coding are more prominent than those with  proximity coding.

\subsection{Relation to existing works}
	Besides different coding schemes for cell analysis, coding of the raw labeled data is also widely adopted for computer vision tasks with natural scenes. The anchor box introduced in YOLO v2 \cite{YOLO_v2} resembles convolution kernels with different shapes \cite{learn_to_count_with_boosting}. The watershed transformation  \cite{deep_watershed} for semantic segmentation is similar to the proposed repel coding. The difference is that in watershed transformation, boundary information is of interest, while for cell detection, center information is the final output. A two-step coding scheme\cite{deep_watershed} inspired by the watershed algorithm is effective in many semantic segmentation applications. The first step involves coding object boundaries. In this step, the coded response at each pixel is a two dimensional unit vector that points to the closest boundary pixel. The coding in the first step aims at maximizing the reversibility of the raw labels. In the second step, the coded response at each pixel is the distance from a pixel to its closest boundary pixel. The second step focuses more on entropy maximization. These two steps are cascaded in the watershed transformation pipeline.

	%We also noticed that this ground truth coding technique is widely used in different computer vision tasks, but a systematic analysis of these coding work is not discussed before. 	
	
	In general, different coding schemes are different ways to transfer an end-to-end training framework to a stepwise implementation. %since the final outputs should be the coordinates of the cell centers. 
	Another way to understand coding the raw dot labels is taking the neural network as a signal processing system. As in the analog domain, designing filters with shape responses such as the unit impulse response is challenging. By coding the raw dot labels, we impair the ideal response by smoothing it, but such smoothing is preferred sometimes because of its easier implementation.
	
\section{Experiments}
\label{sec:expriment}
In this part, different coding schemes are tested for four types of cells and with two CNN architectures. Experimental results show that the proposed repel coding outperforms existing coding schemes both for cell counting and detection tasks. Discussion of the examples from the four types of cells provides some insights into different coding schemes.

\subsection{Datasets}
	\begin{figure}[t]
		\begin{minipage}[t]{0.99\linewidth}
			\centering
			\includegraphics[width = 0.99\linewidth, trim={0cm, 4.3cm, 14cm, 0cm}, clip]{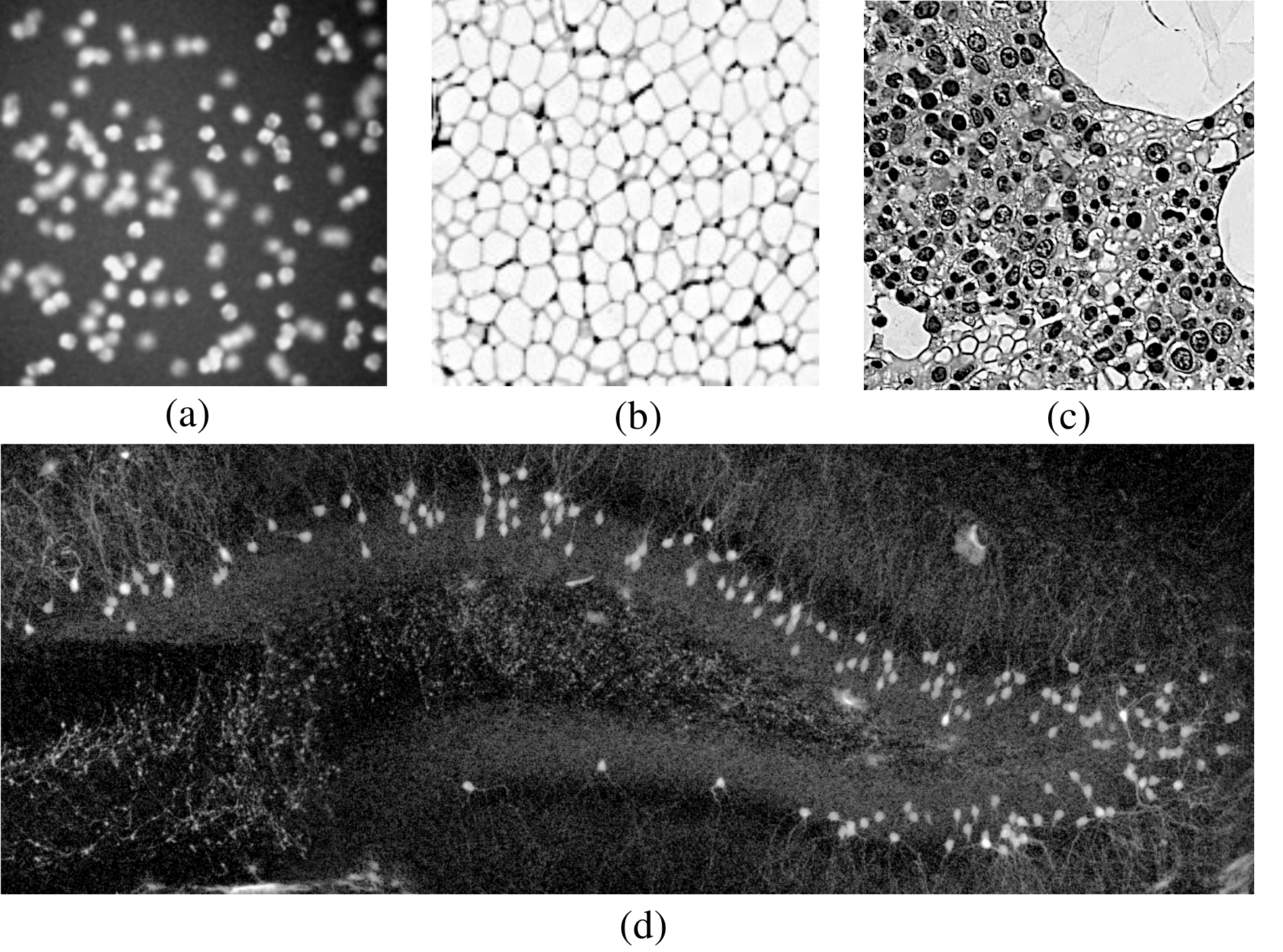}				
		\end{minipage}					
		\caption{Sample images from the four datasets. (a) Synthetic Vgg cells. (b) Adipocyte cells. (c) Human bone marrow cells. (d) Granule cells in the mouse dentate gyrus.}
		\label{fig:dataset_overview}
	\end{figure}
Four datasets are evaluated in the experiments: granule cells in the mouse dentate gyrus (DG), human adipocyte cells (Adip), human bone marrow cells (HBM) and Vgg-generated synthetic cells (Vgg). Fig. \ref{fig:dataset_overview} shows examples from these four datasets. 
\subsubsection{DG dataset}
The DG dataset comprises mouse brain tissues stained by tdTomato after seizure. These brain tissues include the dentate gyrus in the V-shape, and the highlighted cells are granule cells. A Zeiss 780 confocal microscope with a C-Apochromat objective under 10X magnification is used to image the brain tissues. %The near-red excitations are provided by DPSS 561 laser lines, and the emission windows are 571-624 nm. 
The DG dataset contains 26 high resolution dentate gyrute images, and the image size is from $452 \times 942$ to $732 \times 1336$. More details about the DG dataset can be found in our previous work \cite{brain_recon}. 
\subsubsection{Adip dataset}
The Adip dataset\cite{count_ception} contains human subcutaneous adipose tissues obtained from the genotype tissue expression consortium. The available images of Adip dataset are $150 \times 150$, and the size of each single cell is within $32 \times 32$. These adipocyte cells are densely-packed as shown in Fig. \ref{fig:dataset_overview} (b). The Adip dataset contains 200 images.
\subsubsection{HBM dataset}
The HBM dataset\cite{MBM} includes images of healthy human bone marrow from eight different patients. The HBM dataset contains 44 images of size $600 \times 600$, and the objects of interest are the stained nuclei shown in Fig. \ref{fig:dataset_overview} (c). Tissues in HBM dataset are stained with Hematoxylin and Eosin.
\subsubsection{Vgg dataset}
The Vgg dataset\cite{vgg_data} is a group of synthetic cells. With a cell simulation method\cite{cell_simutation_1}, cell detection algorithms can be easily evaluated with different imaging settings, such as cell overlap, out-of-focus blurring, and size variation. The Vgg dataset contains 200 images of size $256\times 256$.

\subsection{Experimental settings}
	Two network architectures are tested in these experiments. The training settings, including the cost function, the learning rate, and the optimization algorithm are kept the same through different experiments. The learning rate is set to $10^{-5}$, the optimizer is Adam with default parameters \cite{Adam}, and the training batch size is set as eight for HBM, Adip and Vgg datasets, and two for the DG dataset, to fit in the 12GB memory of the NVIDA TITAN Xp used in the experiments.
	
	\subsubsection{Cost function}
	To be consistent with the previous works on cell counting and detection\cite{compressive_sensing_counting,learn_to_count_with_boosting,yao_16}, the $L2$ norm is used as the cost function for training,
	\[cost = \|y - y'\|^2_2,\]
	\noindent{where $y$ is the ground truth output coding generated from the raw dot labels, and $y'$ is the output of the CNN. Fig. \ref{fig:coding_overview} illustrates the four coding schemes.}
	
	\subsubsection{Network architectures}
\begin{figure}[t]
		\begin{minipage}[t]{0.99\linewidth}
			\centering
			\includegraphics[width = 0.99\linewidth, trim={0cm, 2.2cm, 10cm, 0cm}, clip]{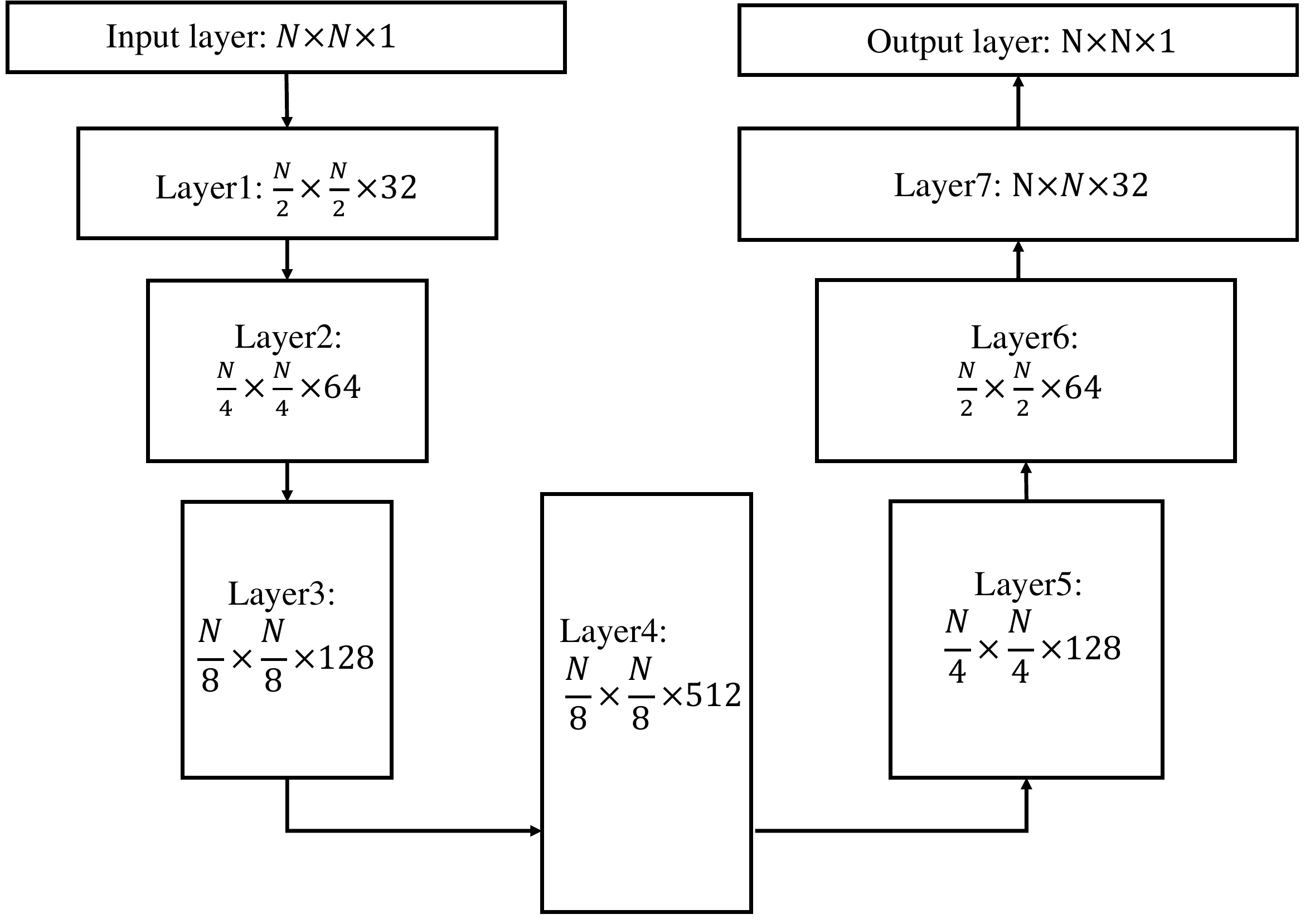}				
		\end{minipage}					
		\caption{The architecture of the CNN in this experiment is based on a U-net with eight layers. If all layers are implemented with plane convolutional blocks, the network is the same as the FCNN-A \cite{xie_16}.}
		\label{fig:network_architecture}
	\end{figure}

	Two CNNs  based on the Unet \cite{Unet} architecture are evaluated in the experiments. One CNN is the same as the FCNN-A \cite{xie_16}, and the other one replaces the convolutional layers in the FCNN-A with residual convolution blocks. The overall architecture of the FCNN-A is shown in Fig. \ref{fig:network_architecture}. The activation function used in all layers is rectified linear unit (ReLU). The receptive fields of these two CNNs are both $38 \times 38$.

	\begin{figure}[t]
		\begin{minipage}[t]{0.99\linewidth}
			\centering
			\includegraphics[width = 0.99\linewidth, trim={0cm, 4cm, 22cm, 0cm}, clip]{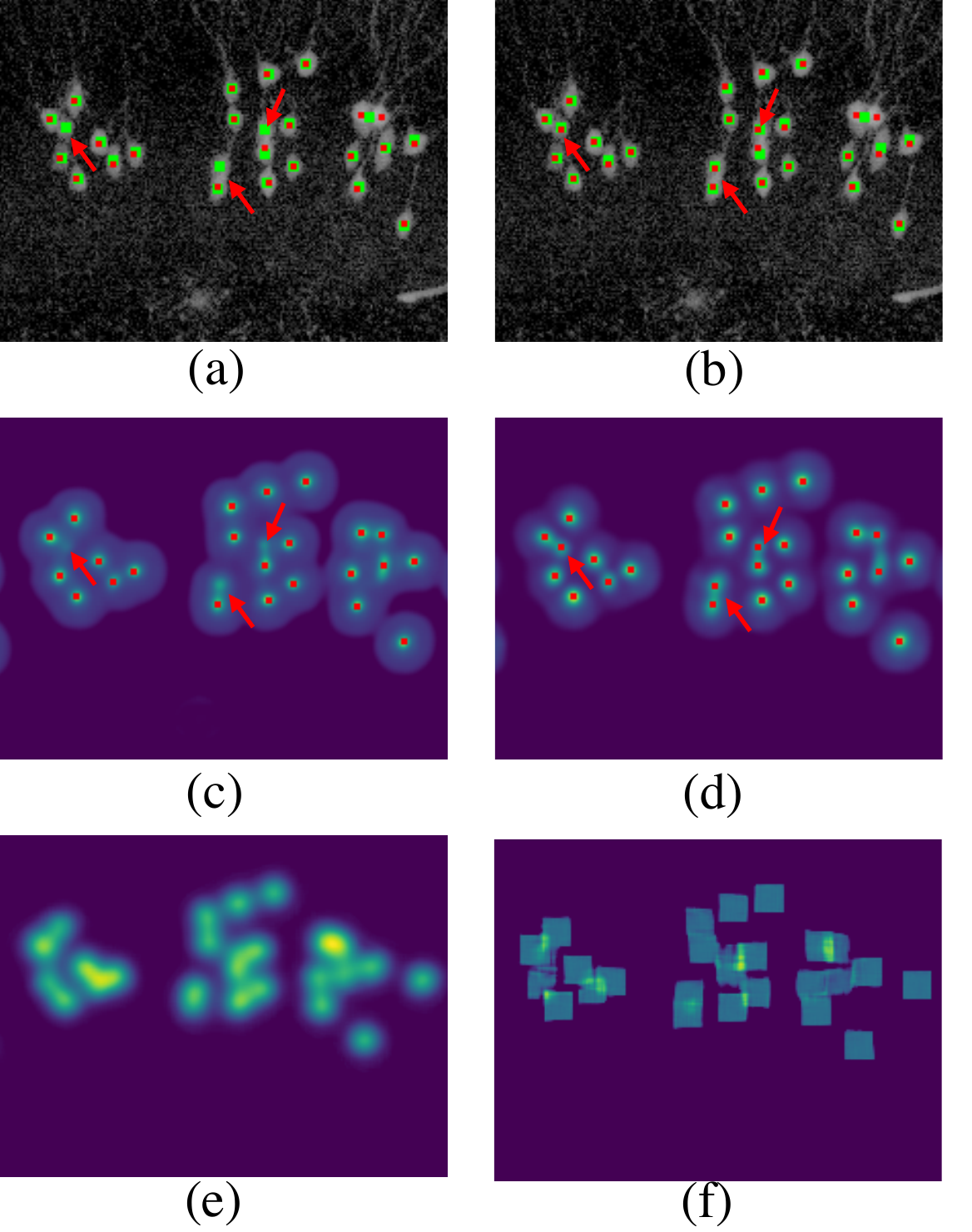}				
		\end{minipage}					
		\caption{(a) Detection results with  proximity coding on an example from the DG dataset. (b) Detection results with the proposed repel coding. (c) The proximity coding output. (d) The repel coding output. (e) The Gaussian kernel coding output. (f) The rectangle kernel coding output. The green dots indicate labeled cell centers, and the red dots represent detected cell centers.}
		\label{fig:result_overview}
	\end{figure}
	
	\begin{figure*}[t]
		\begin{minipage}[t]{0.99\linewidth}
			\centering
			\includegraphics[width = 0.99\linewidth, trim={0cm, 10cm, 15cm, 0cm}, clip]{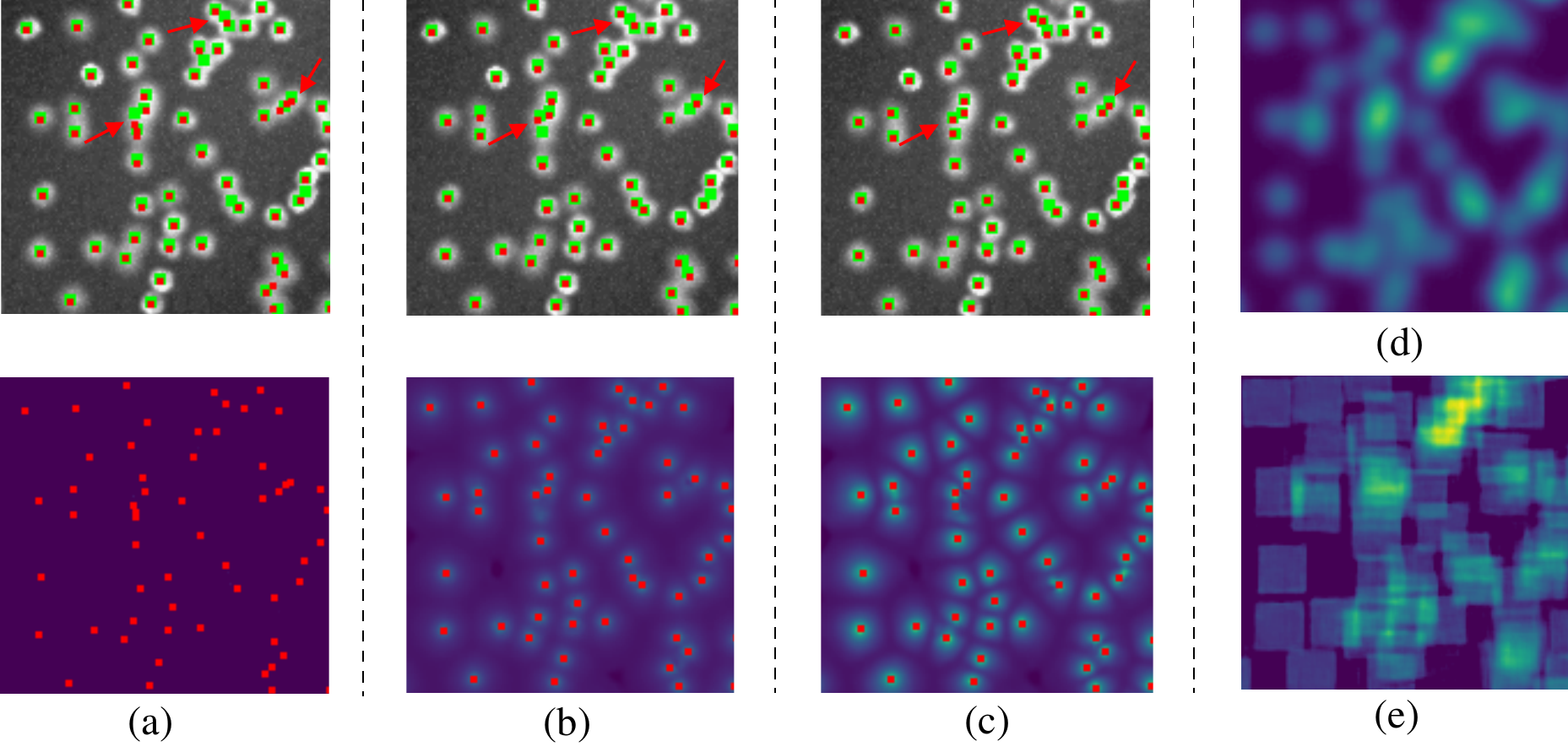}				
		\end{minipage}					
		\caption{Different coding schemes on an example from the Vgg dataset. The green dots indicate labeled cell centers, and the red dots represent detected cell centers. The proposed method distinguishes adjacent cells in most cases. (a) Dot labels. (b) The proximity coding. (c) The repel coding. (d) The Gaussian kernel coding for counting. (e) The rectangle kernel for counting.}
		\label{fig:result_overview2}
	\end{figure*}
	
	\begin{figure*}[t]
		\begin{minipage}[t]{0.99\linewidth}
			\centering
			\includegraphics[width = 0.99\linewidth, trim={0cm, 12.2cm, 18cm, 0cm}, clip]{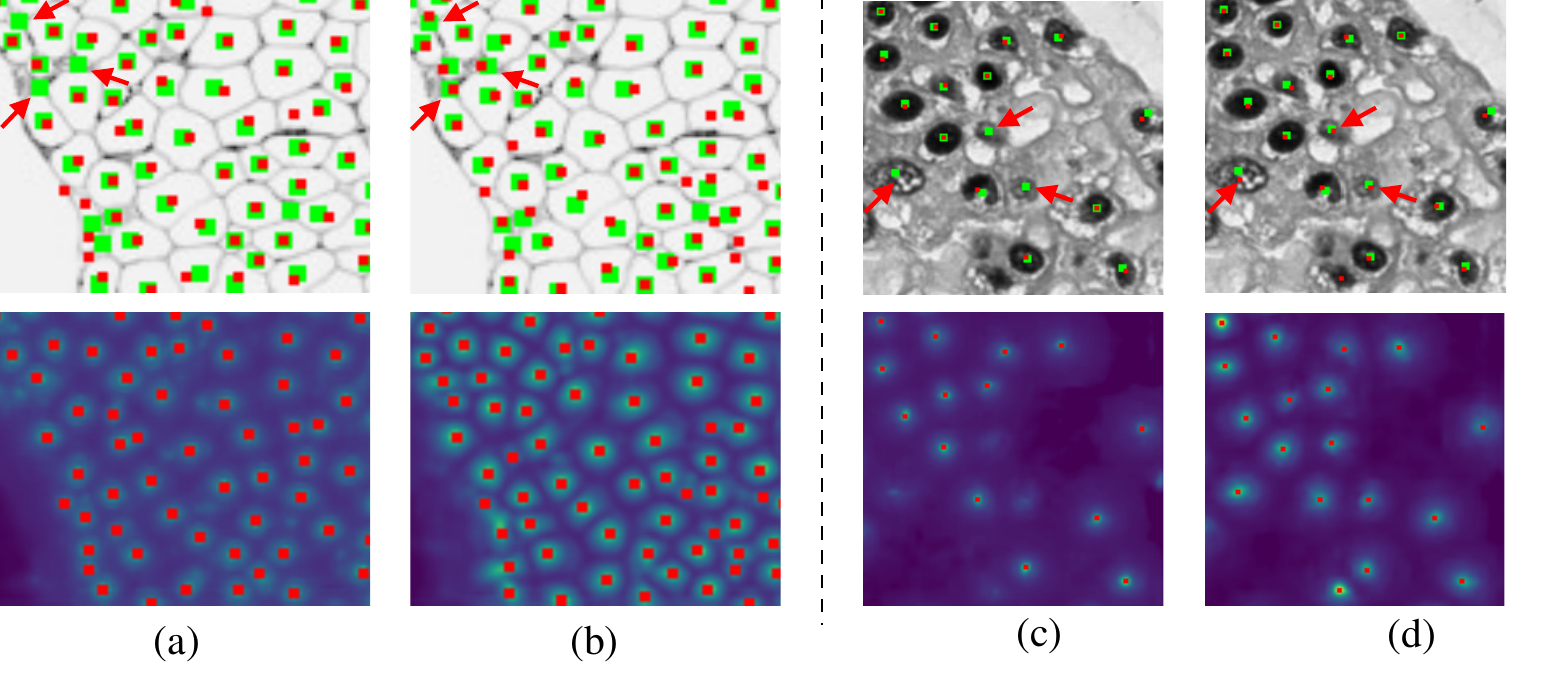}				
		\end{minipage}					
		\caption{(a) and (b) are  proximity and  repel codings of an example from the Adip dataset. (c) and (d) are  proximity and  repel codings of an example from the HBM dataset. The green dots indicate labeled cell centers, and the red dots represent detected cell centers.}
		\label{fig:MBM_Adip}
	\end{figure*}
		
%	\begin{figure*}[t]
%		\begin{minipage}[t]{0.99\linewidth}
%			\centering
%			\includegraphics[width = 0.99\linewidth, trim={0cm, 7cm, 12cm, 0cm}, clip]{figures/Adip.pdf}				
%		\end{minipage}					
%		\caption{Adip}
%		\label{fig:result_overview2_adip}
%	\end{figure*}
	
%	\begin{figure*}[t]
%		\begin{minipage}[t]{0.99\linewidth}
%			\centering
%			\includegraphics[width = 0.99\linewidth, trim={0cm, 10.5cm, 17cm, 0cm}, clip]{figures/MBM.pdf}				
%		\end{minipage}					
%		\caption{MBM}
%		\label{fig:result_overview2_mbm}
%	\end{figure*}
	
	\subsubsection{Evaluation}
	\begin{algorithm}[t]
		\caption{{F1 Score for Evaluation}}
		\label{alg:f1}
		\begin{algorithmic}	
			\State{\hspace{-1em}\textbf{Inputs:}}
			\State{\hspace{-0em} $D_{list}$: List of detected centers}
			\State{\hspace{-0em} $G_{list}$: List of ground truth centers}
			\State{\hspace{-1em}\textbf{Outputs:}}
			\State{\hspace{-0em} $F1$: F1 score}
			\State{}

			\State{\hspace{-1em}\textbf{Initialization:} }
			\State{\hspace{-0em}$1.$ $D_{matched} = [inf, ..., inf]$}				
			\State{\hspace{-0em}$2.$ $G_{matched} = [inf, ..., inf]$}				
			\vspace{0.3em}
			\State{\hspace{-1em}\textbf{Pair Match:} }
			\While {$D_{list}$ is not empty and $G_{list}$ is not empty}
			\State{\hspace{-0em}$3.$ $D_{idx}, G_{idx}, dist = get\_closest\_pair(D_{list}, G_{list})$}				
			\State{\hspace{-0em}$4.$ $D_{matched}(D_{idx}) = dist$} 
			\State{\hspace{-0em}$5.$ $G_{matched}(G_{idx}) = dist$} 
			\vspace{0.3em}
			\EndWhile

			\vspace{0.5em}
			\State{\hspace{-0em}$6.$ $Acc = \frac{\#(D_{matched}\le thresh)}{\#D_{list}}$} 
			\State{\hspace{-0em}$7.$ $Rec = \frac{\#(G_{matched}\le thresh)}{\#G_{list}}$} 
			\State{\hspace{-0em}$8.$ $F1 = \frac{1}{\frac{1}{Acc} + \frac{1}{Rec}}$} 
		\end{algorithmic}
	\end{algorithm}

	Since coding schemes for both counting and detection are compared in the experiments, two measures are adopted here. For codings with the Gaussian kernel and the rectangle kernel, the integration over the outputs is taken as the total number of cells. For  raw dot coding,  proximity coding, and  repel coding, cell centers are extracted by local maximum detection. The F1 score \cite{F1_score} is used as a comprehensive index to evaluate the detection accuracy. Alg. \ref{alg:f1} summarizes the F1 score calculation. In Alg. \ref{alg:f1}, each non-paired detected cell center is matched to the closest non-paired ground truth cell center. Cell centers in the detected list, $D_{list}$, and the ground truth list, $G_{list}$, can be paired only once. A match with the distance less than the average radius of the cell is considered to be a successful match since the cell size variant within a dataset is not much. The average radius is $8$ pixels for the DG dataset, $11$ pixels for the Adip dataset, $15$ pixels for the HBM dataset, and $11$ pixels for the Vgg dataset. 
	
\subsection{Results and analysis}
	With the four datasets, five coding methods, and two CNN implementations, 40 sub-experiments are evaluated. Each sub-experiment tests one coding method with one CNN implementation on one dataset. For each dataset, 80\% percent of the data are used for training, and 20\% are used for testing. Each sub-experiment is run five times with random training/test splitting, and the average performance is reported in Table \ref{table:fcnn_net} and Table \ref{table:res_net}. The proposed repel coding achieves the best performance in most sub-experiments. The only exception is the sub-experiment on the Vgg dataset with the FCNN-A implementation, where the performance of raw dot labels is slightly better than the proposed method. The reason may be that the cell variance in the Vgg dataset is less than in the other datasets, and thus is a less challenging dataset. Comparing the results in Table \ref{table:fcnn_net} and Table \ref{table:res_net}, another observation is that the performance of all the coding methods benefits from the residual convolution blocks. This result is expected since the increased effectiveness of the residual convolution block is demonstrated in previous cell analysis works\cite{yao_16,learn_to_count_with_boosting}.
	
	To clarify why the proposed repel coding outperforms others, examples from four datasets are shown in Figs. \ref{fig:result_overview}-\ref{fig:MBM_Adip}. Fig. \ref{fig:result_overview} shows an example from the DG dataset with the FCNN-A network. When two cells are close,  proximity coding tends to merge the two centers. By comparing the prediction results with the illustrations in Fig. \ref{fig:coding_overview}, we can find the reason. The proposed repel coding suppresses the responses of pixels that lie in the middle of two cell centers, and boosts the responses that are close to cell centers. The outputs from the Gaussian kernel coding and the rectangle kernel coding are as expected in Fig. \ref{fig:result_overview}, and do not have the ability to recover cell centers.  
	Fig. \ref{fig:result_overview2} shows an example from the Vgg dataset trained by the CNN with residual convolutaion blocks. The advantage of the repel coding is obvious in the partially occluded regions. The outputs of the raw dot labels in Fig. \ref{fig:result_overview2} (a) are unstable, and tend to output duplicated cell centers. In addition, we find that training with raw dot labels can easily diverge if the training batch size is less than eight on Adip and Vgg datasets, and this does not happen with the other coding schemes. Image sizes of Adip and Vgg datasets are smaller than those of DG and HBM datasets. This may be due to the sparsity in the raw dot labels that leads to insufficient positive training examples. 
	At last, Fig. \ref{fig:MBM_Adip} compares the performance of proximity coding and repel coding on Adip and HBM datasets. In these two datasets, occlusion is less common, but cell appearance varies more. Because the proposed repel coding has a larger reversibility index, the repel coding generally provides stronger responses around cell centers, resulting in more robust center detection. 
	 %The variation of cell appearance in the Vgg dataset is the least among the four datasets. As a result, all the coding schemes also perform best on Vgg dataset.

	\begin{table}[t]
		\centering
		\caption{F1 score of the FCNN-A}
		\begin{tabular}{ | c | c| c | c | c | c | }
			\hline
			 & Dot label & Gaus. kernel &  Rec. kernel & Proximity & \textbf{Repel}    \\ \hline
			DG & 0.8526 & 0.8431 &  0.8431 & 0.9199 & \textbf{0.92267}    \\ \hline
			Apip & 0.8495 & 0.7918 & 0.7645 & 0.8437 & \textbf{0.8784}    \\ \hline
			HBM & 0.7752 & 0.8685 &  0.8110 & 0.7333 & \textbf{0.8773}    \\ \hline
			Vgg & \textbf{0.9585} & 0.9169 &  0.8728 & 0.9545 & 0.9583    \\ \hline
			
		\end{tabular}
		\label{table:fcnn_net}
	\end{table}
	\begin{table}[t]
		\centering
		\caption{F1 score of the FCNN-A with res. blocks}
		\begin{tabular}{ | c | c| c | c | c | c | }
			\hline
			 & Dot label & Gaus. kernel &  Rec. kernel & Proximity & \textbf{Repel}    \\ \hline
			DG & 0.8887 & 0.8772 &  0.8940 & 0.9282 & \textbf{0.9337}    \\ \hline
			Apip & 0.8764 & 0.8177 & 0.7799 & 0.9028 & \textbf{0.9038}    \\ \hline
			HBM &  0.8370 & 0.8628 &  0.8121 & 0.8933 & \textbf{0.9011}    \\ \hline
			Vgg & 0.9520 & 0.9247 &  0.8992 & 0.9676 & \textbf{0.9695}    \\ \hline
			
		\end{tabular}
		\label{table:res_net}
	\end{table}
	
\begin{comment}

	\begin{table*}[t]
		\centering
		\footnotesize
		\caption{FCNN with plane network}
		\begin{tabular}{| c | c | c | c | c | c | c | }
			\cline{1-7}
			 & \multicolumn{3}{|c|}{Recall} & \multicolumn{3}{|c|}{Precision} \\ \hline
			 & Dot label & Proximity & Repel & Dot label & Proximity & Repel   \\ \hline    		
			 DG & 0.9167 & 0.9110 & 0.9347 & 0.8056 & 0.9373 & 0.9163 \\ \hline 
 			 Apip &0.9127 & 0.7832 & 0.8926 & 0.8096 & 0.9338 & 0.8804 \\ \hline 
 			 HBM & 0.8345 & 0.6052 & 0.8507 & 0.7329 & 0.9479 & 0.9124 \\ \hline 
			 Vgg & 0.9488 & 0.9184 & 0.9233 & 0.9692 & 0.9947 & 0.9970 \\ \hline 
		\end{tabular}
		\label{table:fcnn}
	\end{table*}

	\begin{table*}[t]
		\centering
		\footnotesize
		\caption{FCNN with res-block}
		\begin{tabular}{| c | c | c | c | c | c | c | }
			\cline{1-7}
			& \multicolumn{3}{|c|}{Recall} & \multicolumn{3}{|c|}{Precision} \\ \hline
			& Dot label & Proximity & Repel & Dot label & Proximity & Repel   \\ \hline    		
			DG & 0.9411 & 0.9249 & 0.9363 & 0.8483 & 0.9458 &  0.9356\\ \hline 
			Apip & 0.9053 & 0.9165 & 0.9195 & 0.8650 & 0.9018 & 0.9005 \\ \hline 
			HBM & 0.8482 & 0.8793 & 0.9125 & 0.8377 & 0.9147 &  0.8963\\ \hline 
			Vgg & 0.9398 & 0.9396 & 0.9438 & 0.9683 & 0.9979 & 0.9973 \\ \hline 
		\end{tabular}
		\label{table:res}
	\end{table*}
\end{comment}
	
\section{Conclusion}
\label{sec:conclusion}
	In this paper, after reviewing recent learning-based works on cell counting and detection, the common step of coding raw dot labels is extracted and discussed. Two center coding criteria are proposed: entropy and reversibility. These two criteria help predict the performance of a coding scheme at the training and prediction steps. A new coding scheme, repel coding, is proposed for a better balance with these two center coding criteria. Experimental results verify the effectiveness of repel coding for cell detection on four types of cells. In the future, we would like to explore more about the cell activation topology with the detected cell centers.
	
	%The granule cell detection together with the brain reconstruction identify the 3D coordinates of all the activated cell in the dentate gyrus in mouse brains. With this information, we able to ask questions such as if the activation topology is how spatial memory are coded in the mouse brain.

\section*{Acknowledgements}
The authors wish to thank Dr. Suchitra Joshi for her help preparing the brain tissues of the DG dataset, and  Anvitha Kambham and Smriti Subedi for their help labeling the DG dataset.

\label{sec:ref}
\bibliographystyle{IEEEbib}
\bibliography{refs}
\end{document}